# CROP RECOMMENDATION WITH MACHINE LEARNING: LEVERAGING ENVIRONMENTAL AND ECONOMIC FACTORS FOR OPTIMAL CROP SELECTION

Steven Sam[1] and Silima Marshal D'Abreo [2]
[1, 2] Department of Computer Science
College of Engineering, Design and Physical Science
Brunel University London
steven.sam@brunel.ac.uk

**Abstract**

Agriculture constitutes a primary source of food production, economic growth and employment in India, but the sector is confronted with low farm productivity and yields aggravated by increased pressure on natural resources and adverse climate change variability. Efforts involving green revolution, land irrigations, improved seeds and organic farming have yielded suboptimal outcomes. The adoption of innovative computational solutions such as crop recommendation systems is considered as a new frontier to provide insights and help farmers adapt and address the challenge of low productivity. However, existing agricultural recommendation systems have predominantly focused on environmental factors and narrow geographical coverage in India, resulting in limited and robust predictions of suitable crops with both maximum yields and profits. This work incorporates both environmental and economic factors and 19 crop varieties across 15 states as input parameters to develop and evaluate two recommendation modules – Random Forest (RF) and Support Vector Machines (SVM) – using 10-fold Cross Validation, Time-series Split and Lag Variables approaches. Results show that the 10-fold cross validation approach produced exceptionally high accuracy (RF: 99.96%, SVM: 94.71%), raising concerns of overfitting. However, the introduction of temporal order, which aligns more with real-world scenarios, reduces the model performance (RF: 78.55%, SVM: 71.18%) in the Time-series Split approach. To further increase the model accuracy while maintaining the temporal order, the Lag Variables approach was employed, which resulted in improved performance (RF: 83.62%, SVM: 74.38%) compared to the 10-fold cross validation approach. Overall, the models in the Time-series Split and Lag Variable Approaches offer practical insights by handling temporal dependencies and enhancing its adaptability to changing agricultural conditions over time. Consequently, the study shows the Random Forest model developed based on the Lag Variables as the most preferred algorithm for optimal crop recommendation in the Indian context.

1. Introduction

Agriculture is not only fundamental for food production but also constitutes a primary source for economic growth, employment and improvement of the wellbeing of many people globally. For example, the World Bank reports that agriculture constitutes about 4% of the world's total gross domestic product (GDP), and in certain least developed nations, its contribution to GDP exceeds 25%. In India, agriculture accounts for 17% of total country's GDP making it a critical sector for the country's long-term and inclusive economic growth (IBEF, 2023). The country leads in milk, spices, cotton, and pulses production globally, and it is recognised as the second-largest producer of fruits, vegetables, rice and wheat (IBEF, 2023). India is also among the top exporters of rice, spices and meat (Gulati et al., 2023). However, despite this significant development, farmers in India are confronted with low farm productivity and yields aggravated by increased pressure on natural resources (e.g., soil, air and forest) and insufficient rainfall linked to adverse climate change variability (Ghost, 2019; Gorain et al., 2024). Approximately, only 45% of the areas occupied by 145 million small farm landholdings in India receive proper rainfall suitable for growing crops (Manjunathan et al., 2020). The challenges of low productivity in Indian agriculture has put many people at the risk of experiencing food scarcity and financial difficulties, and farmers who are unable to repay loans taken from banks opt committing suicides (Garanayak et al., 2021). Significant efforts are made by the government of India to address the challenge of low productivity including the green revolution, land reforms, irrigation facilities, improved seeds, organic farming, water management and agriculture infrastructural investment. However, while these initiatives undeniably assist farmers throughout the agricultural cycle, the core challenge of low productivity persists particularly at the grassroots level.

The adoption of innovative computational solutions such as crop recommendation systems is considered as a new frontier to help farmers adapt and address the challenge of low productivity. Crop recommendation system utilises algorithms and diverse data parameters such as weather data, on farm data, soil nutrients and market information to forecast suitable crops with maximum yields and profits (Garanayak et al.,2021; Pudumalar et al., 2017; Kumar et al., 2019). Research shows that crop recommendation systems have been successfully applied to recommend crops to farmers (Geetha et al., 2020), detect plant disease (Shoaib et al., 2023), provide financial help, irrigation facilities and insurance to the farmers' crops (Jaiswal et al., 2020) and maximise crop yield (e.g. Elomda et al., 2014, Patel and Patel, 2020). Most crop recommendation systems incorporate one or more machine learning algorithms to make an efficient and accurate prediction of optimal crop to improve yield and profitability (Dey et al., 2024; Doshi et al., 2018; Lata and Chaudhari, 2019;Patel and Patel and;, 2020). For instance, Geetha et al., (2020) use Random Forest to generate a crop recommendation system with 97% accuracy. Kavita and Mathur (2020) implement a recommendation system to predict a crop with the highest yield using Decision Tree, Linear Regression, Lasso Regression and Ridge Regression. Decision Tree provides a higher accuracy of 98.62% compared to the other algorithms. Lata and Chaudhari (2019) combine Random Forest, J48, Bayes Net and KStar to develop a recommendation system. Random Forest performs the best with an accuracy of 97.89%. In another study by Kumar et al., (2020) Support Vector Machine outperforms Random Forest, Decision Tree Logistic Regression with a highest accuracy of 89.66% while Random Forest gave only 88%. Devi and Selvakumari (2022) designs a recommendation system based on three algorithms: Naïve Bayes, Linear Regression and Random Forest. Random Forest gave 99.09% accuracy compared with 90.9% of Naïve Bayes. Collectively, research into crop recommendation systems is growing but challenges remain to make the recommendation models more robust and provide accurate forecasts to farmers in diverse socio-economic and resource-constrained contexts.

One of the most important challenges in developing accurate crop recommendation models involves selecting relevant input parameters or dataset for training the models. Without adequate, accurate and relevant data, even the most robust machine learning algorithms will not perform well to give the expected results (Patel and Patel, 2020; Rani et al., 2023). Across most of the studies (e.g., Aadithya. et al., 2016; Arooj et al., 2017; Doshi et al.,2018; Elomda et al., 2014; Rani et al., 2023 ), the environmental and economic features are the two broad categories of input parameters used separately for training crop recommendation models. For example, in a study conducted by Elomda et al., (2014) crops are suggested taking into consideration temperature, water and soil as the main features of the model. Jain and Ramesh (2020) developed a crop selection method to maximise crop yield and based on weather conditions and soil parameters like soil pH, soil nutrients and soil type while Patel and Patel (2020) developed their recommendation system considering only soil features like soil types, pH, electric conductivity, organic carbon, nitrogen, phosphorus, sulphur, zinc, boron, iron, manganese and copper. Similarly, few researchers (e.g., Aadithya. et al., 2016; Filippi et al., 2017; Raja et al., 2017) have also used the economic features, which constitute price, market conditions and crop cultivation costs, to develop models that recommend profitable crops to farmers. For instance, Aadithya et al., (2016) developed a crop recommendation model including cost as an economic parameter to recommend the most suitable crop to the farmers. Filippi et al.'s, (2017) recommendation model incorporated economic features such as the availability of tools and tractors, their operating costs and the working speeds along with environmental features like the area available for cultivation, the sequence of operations required to grow each crop and the corresponding time windows to optimise agriculture profits. In the Indian context, most studies have almost exclusively focused on the environmental attributes with limited consideration for the economic parameters (Reddy and Kumar, 2021; Pudumalar et al., 2017; Garanayak et al., 2021).

The objective of this paper is to develop a more comprehensive and inclusive crop recommendation system that incorporates the necessary input parameters for suggesting a wide range of optimal crops to farmers in India. There are approximately 145 million landholdings in India (with an average farm size of 1.08 hectares) upon which diverse crops are cultivated (Manjunathan et al., 2020). The small land size implies that farmers need appropriate knowledge and information to cultivate the right crops that generate high profitable yields in the face of the increasing climate change variability, environmental challenges, market externalities and low productivity. To achieve this, we show that any recommendation system developed needs to be comprehensive and robust to forecast crop choices accurately for farmers across all geographical areas in India (Pudumalar et al., 2017, Reddy and Kumar, 2021; Patel and Patel, 2020). We use eight environmental factors () along with major economic factors related to cost and price: operational cost, fixed cost, total cost of cultivation and minimum support price. In addition, the recommendation system involves 19 crops across 15 states as input parameters in India, extending the work of Pudumalar et al., (2017) and Reddy and Kumar (2021) who have used a narrow range of crops (approximately five to ten crops) within a district or state.

## 2. Method

This section describes the approaches for: (i) data collection and pre-processing, (ii) exploratory data analysis (iii) modelling, and (iv) performance evaluation.

### 2.1 Data Collection and pre-processing

We combined environmental and economic input parameters to develop and evaluate the accuracy of two machine-learning models (Random Forest and Support Vector Machines) for recommending high yield and profitable crops to farmers. The data were collected and merged from three sources: Kaggle[1] for environmental parameters and the India Directorate of Economics and Statistics (DES)[2] and Farmer's Portal[3] for economic parameters of cost and price. Following the work of Rani et al. (2023)

---

[1] https://www.kaggle.com/datasets/vihith12/crop-yield recommendation-dataset

[2] https://eands.dacnet.nic.in/Cost_of_Cultivation.htm

[3] https://farmer.gov.in/mspstatements.aspx

Crop Recommendation with Machine Learning

and other researchers (e.g. Kulkarni et al., 2018, Doshi, et al., 2018 and Kumar, et al., 2020), we consider a broad range of environmental parameters such as temperature, precipitation, wind speed and humidity along with soil attributes such as nitrogen, phosphorous, potassium (NPK) and soil type. These environmental features are important for improving crop production rates and determining soil sustainability, and therefore provide necessary input parameters for crop recommendation systems (Rani et al., 2023). Similarly, we included Operational Cost (e.g. labour, seed, fertilizer cost, etc.), Fixed Cost (e.g. rental, capital, taxes, etc.) and Total Cost of crop cultivation along with the Minimum Support Price (MSP) set by the Government of India for certain agricultural commodities annually as the economic input parameters.

Following data collection, we developed a data dictionary for both the environmental parameters (Tables 1) and economic parameters (Table 2) to aid in pre-processing and analysis and maintain consistency and better understanding of the data.

Table 1: Data dictionary – environmental parameters

| Variable Name | Variable Description | Variable Type |
|---|---|---|
| State Names | State in which the crop is grown | Character |
| District Names | District in which the crop is grown | Character |
| Crop Year | Year in which the crop is grown | Numeric |
| Season Names | Season in which the crop is grown | Character |
| Crop Names | Name of the crop | Character |
| Area | Area in hectares in which the crop is grown | Numeric |
| Temperature | Temperature in the area where the crop is grown | Numeric |
| Wind Speed | Wind speed in the area where the crop is grown | Numeric |
| Precipitation | Precipitation in the area where the crop is grown | Numeric |
| Humidity | Humidity in the are where the crop is grown | Numeric |
| Soil Type | Soil type in the area where the crop is grown | Numeric |
| N | Nitrogen content in the soil where the crop is grown | Numeric |
| P | Phosphorous content in the soil where the crop is grown | Numeric |
| K | Potassium content in the soil where the crop is grown | Numeric |
| Production | Production of crop in tonnes | Numeric |

The pre-processing steps involved merging dataset, dealing with missing values and handling outliers and skewness to ensure that the data is in comprehensive and usable formats for training and testing the models (Tanasa and Trousse, 2004).

**2.1.1 Merging dataset**

The dataset collected from the three sources were merged together in excel based on three common columns – State Names, Crop Year and Crop Names. Along with this, we had four more columns (Operational Cost, Fixed Cost, Total Cost and MSP). As shown in Table 2, a new variable – 'Yield' was computed by dividing the variable Production by Area (Yield = Production / Area) to incorporate both these variables into a single variable and added to the final dataset. However, before combining the datasets, basic cleaning was done in excel to correct inconsistencies in the data like typos, inconsistent spellings and data entry errors to maintain uniformity. It was essential that the spellings of the values in the variables - Crop Names and State Names are consistent across both the datasets as these were the common columns required to merge two datasets. For example, the state Chhattisgarh was spelled 'Chhattisgarh' in the data collected from Kaggle and 'Chattisgarh' in the data collected from DES. In addition, the crop Bajra was incorrectly written as 'Bajar' in a few instances indicating a typo, which was also corrected. Additionally, we noticed that the years 2011-15 were common crop years for both the datasets but the crop year 2015 had entries of only one State – Odisha, and was also removed from the dataset to avoid data bias.

Table 2: Data dictionary – economic parameters

| Variable Name | Variable Description | Variable Type |
| --- | --- | --- |
| Yield | Yield of the crop cultivated | Numeric |
| Operational Cost | Operational cost of cultivation in Rs. per hectare | Numeric |
| Fixed Cost | Fixed cost of cultivation in Rs. per hectare | Numeric |
| Total Cost | Sum of operational cost and fixed of cultivation in Rs. per hectare | Numeric |
| MSP | Minimum support price fixed by the government | Numeric |

The combined dataset from all the three sources was reduced to 12,839 instances with 20 columns due to the limitation of R Studio to handle large dataset and slow down analysis with numerous features (Shah et al., 2022). It consists of 19 different crops across 15 states in India for the years 2011-14 (common years) which were incorporated in the crop recommendation model. The 19 crops are - Arhar, Bajra, Barley, Cotton, Gram, Groundnut, Jowar, Jute, Maize, Moong, Paddy, Ragi, Rapeseed and Mustard, Safflower, Sesamum, Soyabean, Sunflower, Urad and Wheat. The 15 states are - Andhra Pradesh, Assam, Bihar, Chhattisgarh, Gujarat, Haryana, Karnataka, Madhya Pradesh, Maharashtra, Odisha, Punjab, Tamil Nadu, Uttar Pradesh, Uttarakhand and West Bengal.

### 2.1.2 Handling Missing Values, Outliers and skewness

The feature 'Production' consisted of 89 missing instances randomly with no specific patterns. This made it difficult to compute any measure of central tendency. Additionally, it consisted of only a tiny portion of the total dataset (0.6%) and, therefore, were deleted from the dataset, as generally, if less than 5% of values are missing then it is acceptable to ignore them (Schafer and Graham, 2002). There were no duplicated instances in the dataset that needed cleaning. However, a boxplot analysis coupled with the summary statistics confirmed the presence of outliers in all the numerical variables except MSP. It was unclear if the outliers were implausible values or in fact correctly recorded instances that deviated from the other values in the dataset, as the outliers were a significant part of the dataset. Thus, considering removing outliers could lead to potential loss of important instances, introduce bias and reduce the accuracy of the models, they were not discarded.

Histograms and skewness results revealed some amount of skewness present in all the variables but the variable 'Yield' had extremely high positive skewness. This implies that the majority of instances have very low yields, while a few instances show exceptionally high yields. Given that the skewness of the variable 'Yield' highly deviated from the rest of the values and seemed erroneous, it was deleted from the dataset. This reduced the skewness as shown in Figure 2.

```
Skewness of Temperature: -0.31
Skewness of Wind.Speed: 1.38
Skewness of Precipitation: 9.24
Skewness of Humidity: 0.92
Skewness of N: 3.72
Skewness of P: 2.84
Skewness of K: 2.32
Skewness of Production: 6.12
Skewness of Area: 4.40
Skewness of Yield: 38.58
Skewness of Operational.Cost: 1.24
Skewness of Fixed.Costs: 1.11
Skewness of Total.Cost: 1.00
Skewness of MSP: 0.19
```

```
Skewness of Temperature after removal: -0.31
Skewness of Wind.Speed after removal: 1.38
Skewness of Precipitation after removal: 9.24
Skewness of Humidity after removal: 0.92
Skewness of N after removal: 3.72
Skewness of P after removal: 2.84
Skewness of K after removal: 2.32
Skewness of Production after removal: 6.12
Skewness of Area after removal: 4.40
Skewness of Yield after removal: 8.97
Skewness of Operational.Cost after removal: 1.24
Skewness of Fixed.Costs after removal: 1.11
Skewness of Total.Cost after removal: 1.00
Skewness of MSP after removal: 0.19
```

Figure 1: Skewness before removal        Figure 2: Skewness after removal

In all, following data cleaning, the size of the dataset reduced to 12749 instances and 20 variables.

## 2.2 Exploratory Data Analysis

Exploratory Data Analysis (EDA) is an initial crucial step in the data analysis process that offers a systematic examination and visualisation of the data to gain insights, discover patterns, and uncover relationships among variables (Sahoo et al., 2019). In this study, we carried out EDA in RStudio to explore the connections among multiple variables through correlation analysis, assessment of multicollinearity, and graphical examination. The objective was to identify key variables crucial for constructing the crop recommendation model and to eradicate less significant variables that could potentially lead to modelling issues. Detailed description and result of the EDA are presented in the result section below.

## 2.3 Modelling

We used R studio in this study to develop, train and validate Support Vector Machines and Random Forest and algorithms for accurately recommending optimal crops to farmers. The models were trained and tested by splitting the dataset into 80% training and 20% testing data (Gholamy et al., 2018). This section below describes the algorithms and approaches used to develop the recommendation models, and the resulting accuracies are presented in the result section of the paper.

### 2.3.1 Machine learning models

As observed in literature, Support Vector Machines and Random Forest offer complementary strength for crop recommendation systems (e.g. Geetha et al., 2020; Doshi et al., 2018; Dighe et al., 2018; Bondre and Mahagaonkar, 2019). SVM can handle both linear and non-linear data through kernel functions. They are particularly effective in high-dimensional spaces, making them suitable for complex datasets and mitigating the risk of overfitting due to their focus on maximising the margin between classes (Geetha et al., 2020). In the context of crop recommendation, SVMs can be applied to capture complex interactions between factors such as climate, soil type, and crop attributes (Dighe et al., 2018). However, computational complexity can become an issue, particularly with large datasets. And SVM performance might degrade when dealing with noisy or overlapping classes, which is a common occurrence in real-world datasets.

In contrast, RF models excel in their robustness and ability to handle large datasets without overfitting. By aggregating multiple decision trees, they can provide high predictive accuracy and mitigate individual tree biases (Geetha et al., 2020). It can handle missing values, and its insensitivity to irrelevant features further enhances its suitability for real-world applications. The model's feature importance metrics also aid in selecting relevant variables for crop recommendation systems (Geetha et al., 2020; Doshi et al., 2018). Their adaptability to both classification and regression tasks is valuable in addressing the multifaceted aspects of agricultural data, ultimately leading to improved recommendation accuracy. However, the ensemble nature of RF models can make them less interpretable compared to individual decision trees and it might exhibit a bias towards categorical variables with more levels, affecting the accuracy of variable importance calculations (Varghese, 2018). Further, the dataset's inherent characteristics, such as feature relationships and noise levels influence which model performs better. In all, considering the advantages and disadvantages of both the SVM and RF algorithms, we used them to build recommendation models, compared and evaluated the accuracy for forecast high yield and profitable crops for farmers.

### 2.3.2   Three Approaches used

The two algorithms –SVM and RF were trained using three approaches namely:  10-fold Cross Validation Approach (Approach 1), Time-series Split Approach (Approach 2) and Lag Variables Approach (Approach 3). **Approach 1** builds the models using 10-fold cross validation technique and randomly creates the train and split dataset as generated by the software. Both the training and testing dataset consisted of randomly shuffled selected instances from 2011 to 2014. In k-fold cross-validation, the dataset was partitioned into k subsets and the model was trained and assessed k times, with each fold serving as the validation set in turn. The performance metrics from each fold were averaged, offering an estimate of the model's overall effectiveness and reducing overfitting risk (Pandian, 2023).  However, since the dataset used for the study has observations from the years 2011 to 2014 across five seasons – kharif, rabi, autumn, summer and winter – it was logical to treat the dataset as time-series data. Time series data is a collection of observations obtained through repeated measurements over time (Hayes, 2022). Therefore, it was crucial to build models maintaining the temporal order of the data and capture the fluctuations that occur in the variables over time.

In Time-series Split Approach 2, the observations were arranged in ascending order of year – 2011, 2012, 2013, 2014 and, within each year, instances were arranged based on seasons in the order – winter, summer, kharif, autumn, rabi and whole year (6 categories of the variable Season in the dataset). Then, 80-20% train and test split were done in a way that the developed models were trained using the data from past years (2011-14) and the test data mostly included instances from the year 2014. The split is a standard approach in time series analysis. It aims to stimulate how well the model would perform in predicting future values based on patterns observed in historical data while avoiding the pitfalls of data leakage and ensuring that the temporal order is preserved (Talagala et al., 2018).

In the Lag variables Approach 3, along with the train and test split mentioned in Approach 2, lag features were created for Temperature, Precipitation, Humidity and Wind Speed based on the variable Seasons and for Operational Cost, Total Cost and for MSP based on the variable Year. This is because it is revealed through the EDA process that environmental variables in the dataset tend to fluctuate

Crop Recommendation with Machine Learningwith the seasons while economic variables with the years (more details in section 4.1). There is an evidently increasing pattern in cost and price variables over years, which is obvious considering the economic inflation every year. Creating lag variables is a fundamental technique in time series analysis that involves incorporating past values of a variable as features in a predictive model. These lagged variables capture temporal dependencies and patterns, allowing the model to consider how the variable's history influences its future values and are crucial for addressing autocorrelation, accounting for time delays in effects, and improving predictive accuracy (Shumway and Stoffer, 2011).

## 2.4 Performance Evaluation

The developed models were evaluated with the help of accuracy value, Kappa statistics and F1 score. Accuracy measures the percentage of correct recommendations made by the model. Many studies have used accuracy as an evaluation metric (e.g. Arooj et al., 2018).

$$\frac{\text{True Positives} + \text{True Negatives}}{\text{True Positives} + \text{True Negatives} + \text{False Positives} + \text{False Negatives}} = \frac{\text{N. of Correct Predictions}}{\text{N. of all Predictions}} = \frac{\text{N. of Correct Predictions}}{\text{Size of Dataset}}$$

Figure 3: Accuracy Formula

However, the Kappa statistic (Cohen's Kappa) is most of the times preferred over accuracy because it accounts for chance agreement between predicted and actual outcomes. It is less sensitive to class imbalances and provides a more balanced measure of model performance, helping to avoid misleading assessments and hence will be used as one of the performance metrics. Kappa values range from -1 (no agreement) to 1 (perfect agreement) (Shmueli, 2021).

Additionally, accuracy often disregards the particular error types committed by the model and instead emphasizes overall correctness. It prioritizes a generalized correctness. To assess the model's competence in recognising and predicting True Positives, it is more appropriate to gauge F1 score that combines both precision and recall (EvidentlyAI, n.d). The F1 score is often referred to as the harmonic mean of precision and recall metrics and therefore is a single metric that weighs two ratios (Kranstren, 2020).

$$2 * \frac{\text{Precision} * \text{Recall}}{\text{Precision} + \text{Recall}}$$

Figure 4: F1 score Formula

Precision measures the proportion of correct positive recommendations (true positivies) out of all positive recommendations made by the model (Zhu, 2023).

$$\frac{\text{True Positives}}{\text{True Positives + False Positives}} = \frac{\text{N. of Correctly Predicted Positive Instances}}{\text{N. of Total Positive Predictions you Made}}$$

Figure 5: Precision Formula

Recall measures the proportion of correct positive recommendations out of all true positive instances in the data (Zhu, 2023).

$$\frac{\text{True Positives}}{\text{True Positives + False Negatives}} = \frac{\text{N. of Correctly Predicted Positive Instances}}{\text{N. of Total Positive Instances in the Dataset}}$$

Figure 6: Recall Formula

Hence, accuracy, kappa statistics and F1 scores were compared for the developed models through all the three approaches to recommend the most suitable modelling technique to the farmers to fulfil the research objective.

**Results**

**3.1 Exploratory Data Analysis Results**

**3.1.1 Scatter Plot Analysis**

The scatter plot analysis discovered a significant positive relationship between five pairs of variables - Temperature and Precipitation, Operational Cost and Total Cost, Total Cost and Fixed Cost, K and P and Production and Area. The temperature and precipitation exhibit a negative relationship while all other pairs depict strong positive relationship (Figures 7a, b, c, d and e).

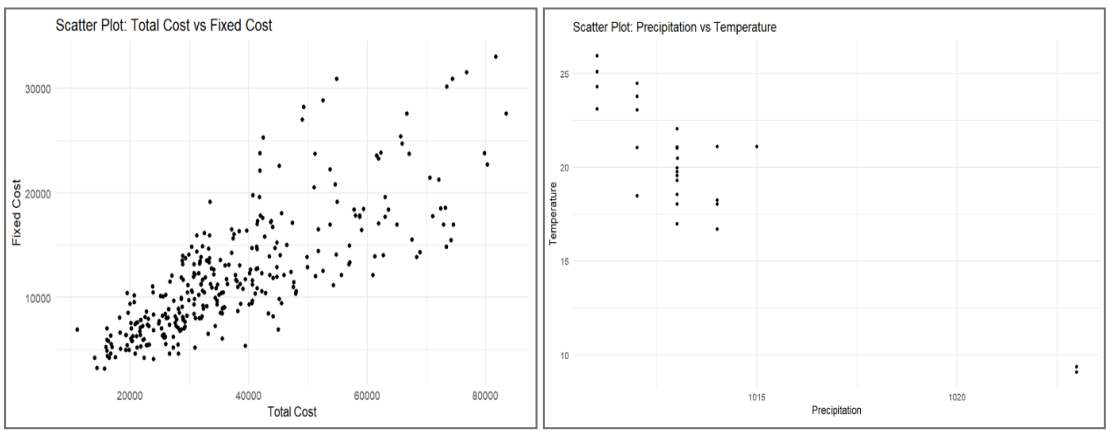

Figure 7a: Total Cost vs Fixed Cost          Figure 7b: Precipitation vs Temperature

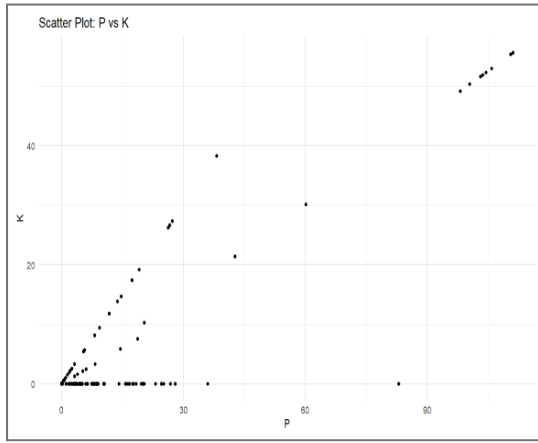 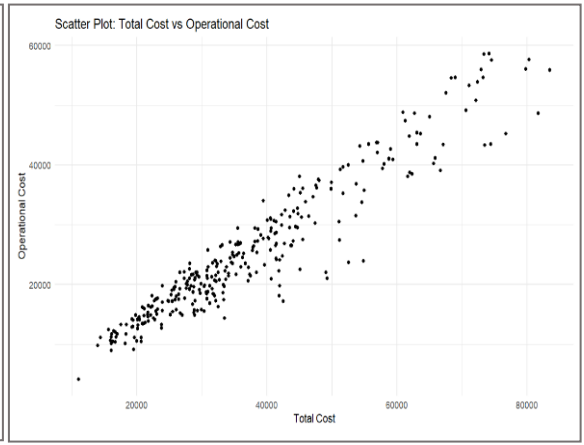

Figure 7c:  P vs K                                            Figure 7d: Total Cost vs Operational Cost

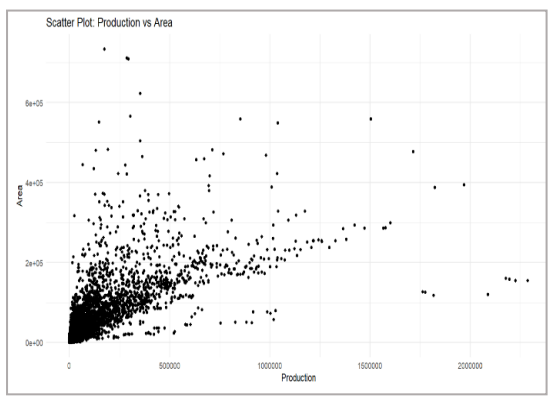

Figure 7e: Production vs Area

The relationship between the variables can indicate the presence of multicollinearity. Multicollinearity in machine learning algorithms can cause interpretability issues that might lead to the loss of reliability in determining the effects of individual features in the model (Bhandari, 2024). Hence, to confirm the presence of multicollinearity and mitigate the risk, we carried out correlation and multicollinearity analysis.

### 3.1.2    Correlation and Multicollinearity Analysis

Results in Figures 8a and 8b show significant correlations between the five pairs of variables associated with crop recommendation prediction. The Temperature and Precipitation variables shared a strong negative correlation of -0.85 while the Total Cost and Operational Cost and Total Cost and Fixed Cost features illustrate strong positive correlation of 0.95 and 0.76 respectively. K and P also have a strong positive correlation of 0.87 followed by Production and Area with a positive correlation of 0.74. Arguably, the significant correlations among multiple variables suggest the presence of multicollinearity between the variables, which can skew the results of the models (Shrestha, 2020). To confirm this, we carried out a multicollinearity test using the Variance Inflation Factor (VIF) technique.

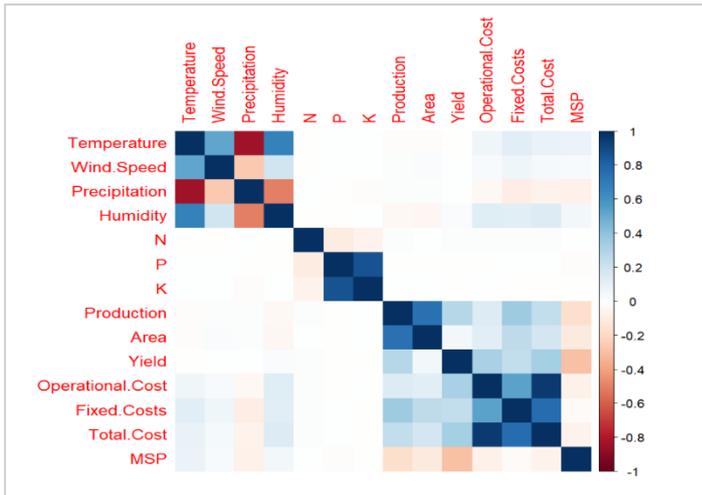
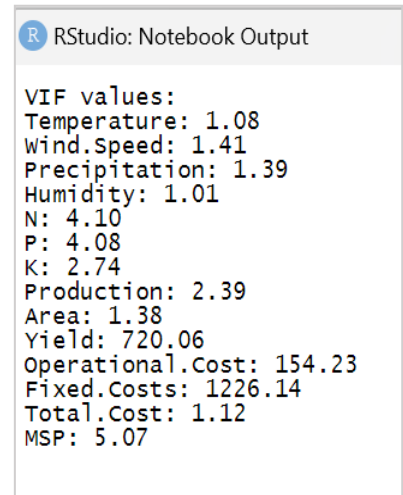

Figure 8a: Correlation Plot    Figure 8c: VIF values

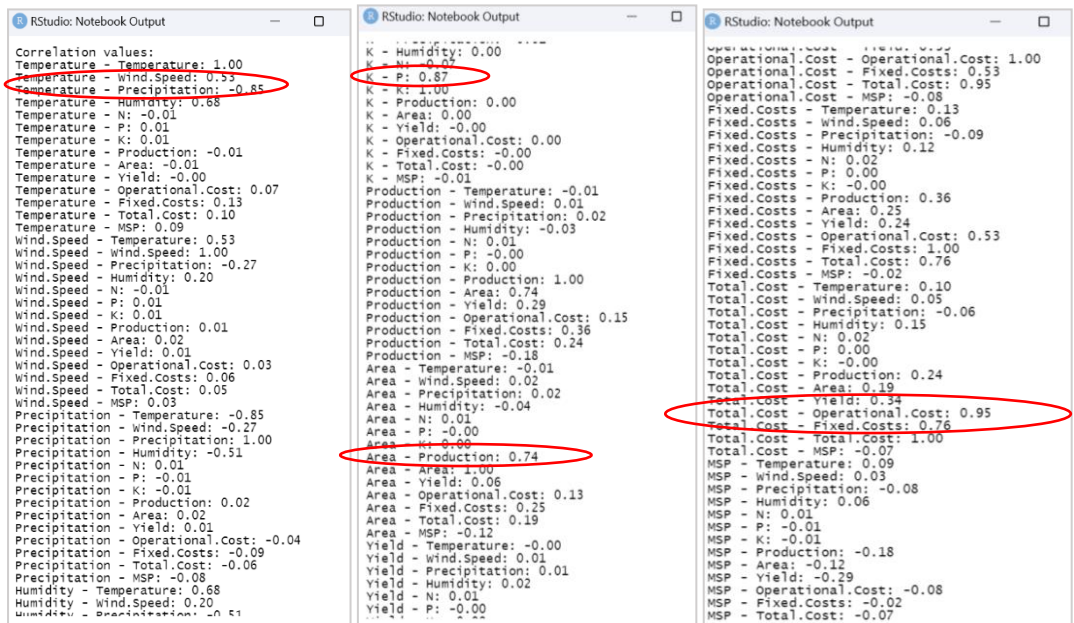

Figure 8b: Correlation Values

The Variance Inflation Factor test evaluates multicollinearity in multiple linear regression models by measuring the extent to which the differing variance of the estimated regression coefficient is inflated (Shrestha, 2020). VIF values < 5 suggests moderate multicollinearity while the VIF values ≥ 5 to 10 and above implies multicollinearity exists among the variables. This can be estimated as:

$$VIF = \frac{1}{1 - R^2} = \frac{1}{Tolerance}$$

In our study, a regression model was designed in RStudio using the numerical variables to test multicollinearity amongst them. The result demonstrates that while some correlation exists, it is not severe enough to cause significant issues in terms of parameter estimation and model performance. The values of VIF all variables are 1< VIF ≤ 5 except for the Yield (720.06), Operational Cost (154.23) and Fixed Cost (1226.14) variables (Figure 8c). The high values suggest that there is a problem of collinearity with these three variables. However, based on the correlation values in Figures 8a and 8b, the Yield variable does not have any strong correlations with other variables but still exhibits high VIF value. This might be attributed to the fact that Yield is derived from two variables, namely Production and Area, which are strongly correlated (with a correlation value of 0.74).

### 3.1.3 Bar Plot Analysis

In the lag variables approach, it was essential to identify how different environmental and economic variables fluctuate with time-series variables – Years and Seasons. We used the bar plot analysis to capture these fluctuations. According to National Geographic (2023), each season has its own light, temperature and weather patterns, and these patterns repeat yearly. Hence, to understand how the environmental variables fluctuate with seasons, standard deviation barplots were plotted for each of the environmental variables - Temperature, Humidity, Wind Speed, and Precipitation against the variable – Season (Figures 9a, b, c and d). A careful observation of all these figures indicate that the bar heights of all these variables fluctuate with every season indicating seasonal fluctuations of environmental variables.

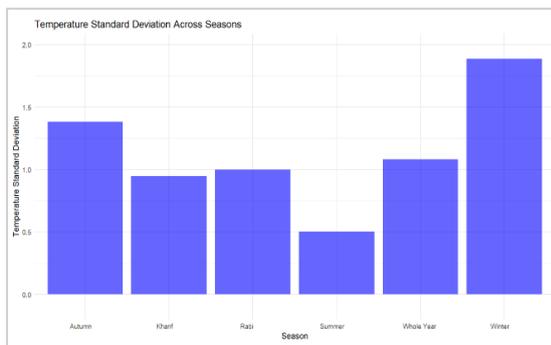
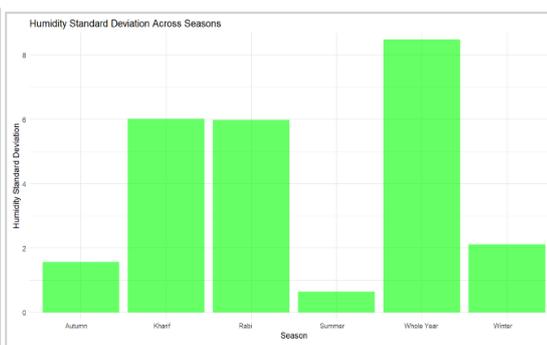

Figure 9a : Temperature Fluctuations with Seasons

Figure 9b : Humidity Fluctuations with Seasons

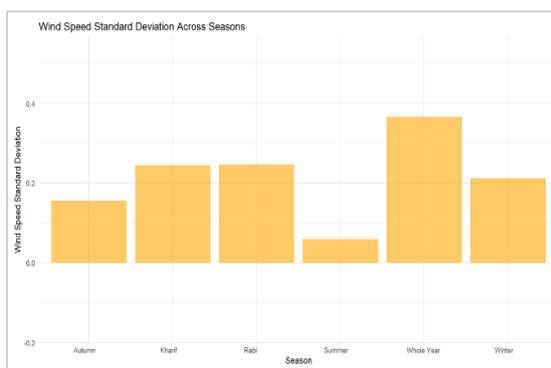
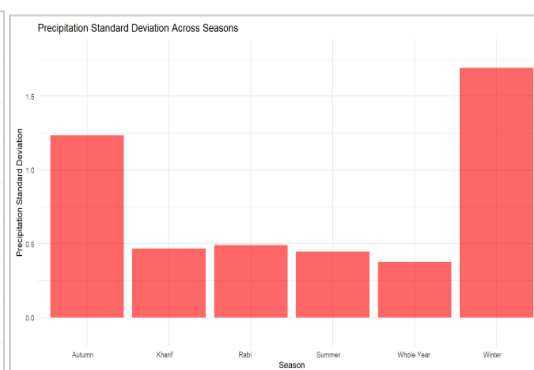

Figure 9c : Wind Speed Fluctuations Fluctuations with Seasons

Figure 9d : Precipitation with Seasons

Similarly, to understand the inflationary trends of economic variables yearly, mean barplots of economic variables – MSP, Operational Cost and Fixed Cost were plotted against each year (Figures 9e, f, g) and as seen, all the economic variables exhibit a yearly increasing pattern.

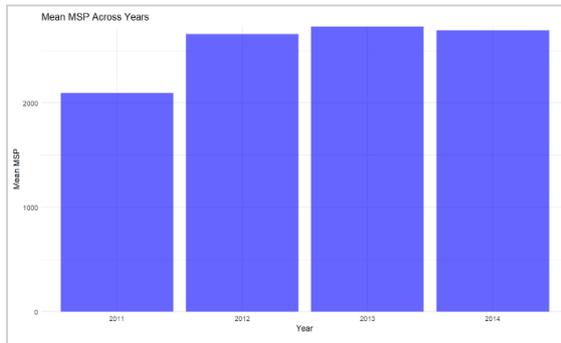
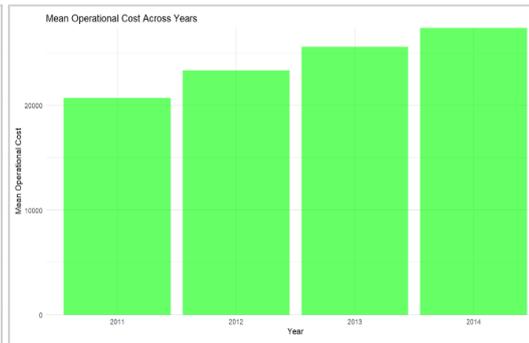

Figure 9e : MSP increase with Years

Figure 8f : Operational Cost increase with Years

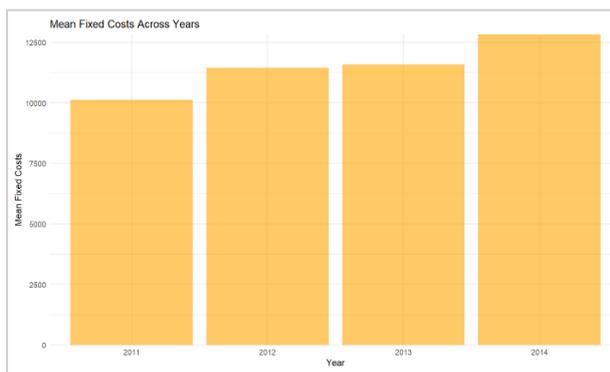

Figure 9g : Fixed Cost increase with Years

In summary, the EDA results with the help of multicollinearity, correlation and scatter plot analysis identified the important variables for modelling and aimed to pinpoint highly correlated variables for tackling multicollinearity issues. Similarly, the barplot analysis established a relationship between multiple variables for successful formation of lag features.

### 3.2 Modelling Results

As mentioned in the methodology section, we used Support Vector Machines and Random Forest to develop the crop recommendation models using RStudio with three approaches – 10-fold Cross Validation Approach, Time-series Split Approach and Lag Variables Approach. The results of the three approaches are described as follow:

### 3.2.1    10-fold Cross Validation Approach (Approach 1)

The essence of the ten-fold cross-validation approach was to evaluate the performance of the models using the train and test split data. The dataset was first randomly shuffled before creating the train and test data. From the result, it is observed that RF performs better than SVM, although both exhibit high performance accuracy above 90% (Table 2).

Crop Recommendation with Machine Learning

Table 2: Performance metrics of 10-fold CV Approach

| Models | Accuracy | Kappa |
|---|---|---|
| Random Forest | 0.9966 | 0.9966 |
| Support Vector Machine | 0.9573 | 0.9538 |

Figure 8 displays the performance metrics of F1-score for each crop. The F1-score close to 1 indicates a strong performance with a good balance between precision and recall. Thus, as can be seen from Figure 10, the RF model can make accurate recommendations for the several crop classes compare with SVM. RF displays approximately the performance metrics of 1 F1-score for all crops.

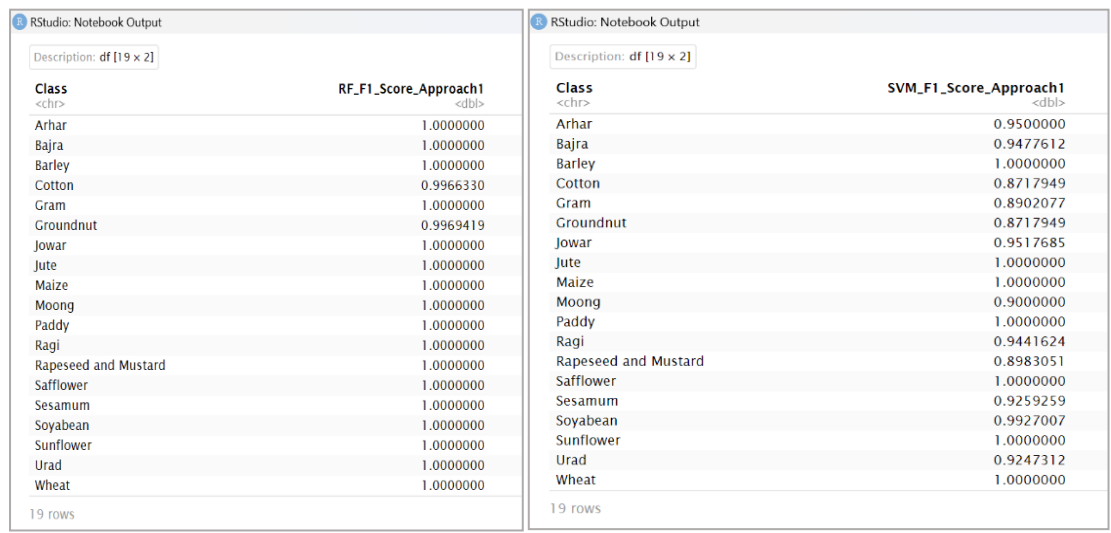

Figure 10: Approach 1 F1 scores for RF and SVM models

In sum, the results of the 10-fold cross-validation shows that the RF model exhibits remarkably high performance across the metrics including accuracy, kappa and F1-scores. However, given that accuracy and kappa statistics align closely in value raises a concern about potential overfitting. It is likely that the model may have captured noise in the training dataset and might not generalise well to new data or unseen data (Priyadharshini et al., 2021). To be confident about the given accuracy, it was essential to evaluate the performance of all these models on a completely new or unseen dataset as described in Approach 2.

### 3.2.2 Time-Series Split Approach (Approach 2)

As outlined in the methodology, this approach arranged the observations in the chronological order of time and then partitioned the dataset into an 80% training and 20% testing ratio while maintaining the chronological sequence of instances, a common practice when handling time-series data to uphold the temporal coherence. Additionally, this partitioning ensured that the model is trained on historical observations (years 2011-13) while forecasting is done on completely unseen instances (year 2014).

```
Overall Statistics
                                          
               Accuracy : 0.7855           
                 95% CI : (0.769, 0.8013)  
    No Information Rate : 0.2098           
    P-Value [Acc > NIR] : < 2.2e-16        
                                          
                  Kappa : 0.764            
                                          
 Mcnemar's Test P-Value : NA               
                                          
Statistics by Class:                       
                                          
                     Class: Arhar Class: Baj
Sensitivity                1.00000  0.2125 
Specificity                0.96701  1.0000 
Pos Pred Value             0.66239  1.0000 
Neg Pred Value             1.00000  0.9751 
Prevalence                 0.06078  0.0313 
```

```
Overall Statistics
                                          
               Accuracy : 0.7118           
                 95% CI : (0.6938, 0.7293) 
    No Information Rate : 0.2098           
    P-Value [Acc > NIR] : < 2.2e-16        
                                          
                  Kappa : 0.6886           
                                          
 Mcnemar's Test P-Value : NA               
                                          
Statistics by Class:                       
                                          
                     Class: Arhar Class: Baj
Sensitivity                0.87742  0.72  
Specificity                0.99248  0.93  
Pos Pred Value             0.88312  0.258 
Neg Pred Value             0.99207  0.990 
Prevalence                 0.06078  0.03  
```

Figure 11a: RF results Approach 2     Figure 11b: SVM results Approach 2

| Class | RF_F1_Score_Approach2 | | Class | SVM_F1_Score_Approach2 |
|---|---|---|---|---|
| Arhar | 0.77500000 | | Arhar | 0.87459807 |
| Bajra | 0.35051546 | | Bajra | 0.34666667 |
| Barley | 0.05263158 | | Barley | 0.07792208 |
| Cotton | 0.96039604 | | Cotton | 0.71895425 |
| Gram | 0.64220183 | | Gram | 0.71337580 |
| Groundnut | 0.98039216 | | Groundnut | 0.77536232 |
| Jowar | 0.69444444 | | Jowar | 0.76190476 |
| Jute | 0.88888889 | | Jute | 0.95238095 |
| Maize | 0.92446043 | | Maize | 0.72374429 |
| Moong | 0.99658703 | | Moong | 0.60000000 |
| Paddy | 0.98765432 | | Paddy | 0.98765432 |
| Ragi | 0.81428571 | | Ragi | 0.73043478 |
| Rapeseed and Mustard | 0.73846154 | | Rapeseed and Mustard | 0.85479452 |
| Safflower | 0.57777778 | | Safflower | 0.57777778 |
| Sesamum | 0.56737589 | | Sesamum | 0.46721311 |
| Soyabean | 0.88000000 | | Soyabean | 0.94642857 |
| Sunflower | 0.78014184 | | Sunflower | 0.75177305 |
| Urad | 0.33451957 | | Urad | 0.66945607 |
| Wheat | 0.75095785 | | Wheat | 0.70397112 |

Figure 11c: Approach 2 F1 scores for RF and SVM models

Surprisingly, when the models are tested on completely unseen data in Approach 2, the accuracy, kappa statistics and F1 scores drop significantly for both the models as shown in Figures 11a, b and c. The drop in these values can be seen to provide a more realistic evaluation of the model's performance compared to the previous approach. In real-world scenarios, models need to be able to generalise to new, unseen data. If the model's performance drops because it struggles to adapt to temporal changes or new patterns in test data, it indicates that the model is not overly optimistic in its recommendation. This drawback necessitated for more accurate model adaptation and retraining strategies to better capture temporal patterns and shifts and is discussed in Approach 3.

### 3.2.3 Lag Variables Approach (Approach 3)

Lag variables, also refer to time lag features, play a crucial role in capturing the historical patterns and temporal dependencies within the data through modelling relationships between past values and observations such as seasonality and long-term trends (Shumway and Stoffer, 2010). In our study, this helps uncover recurring patterns and overarching trends that are critical for accurate crop recommendation. The barplot analysis in Figure12a confirms that environmental variables – Temperature, Precipitation, Humidity and Wind Speed – fluctuate seasonally. Hence, we created lag variables for these variables based on seasons. Similarly, lag variables were also created based on Year

Crop Recommendation with Machine Learning

for economic variables – Operational Cost, Fixed Cost and for MSP – as the variables fluctuate with years.

```r
library(dplyr)

# The dataset is named "cropasc"

# Number of lags you want to create for year
year_lags <- 5

# Columns for which you want to create year lag variables
year_lag_columns <- c("Operational.Cost", "Fixed.Costs", "MSP")

# Create year lag variables for the specified columns
year_lagged_cropasc <- cropasc %>%
  group_by(State, Crop) %>%
  mutate(
    across(all_of(year_lag_columns), ~lag(., order_by = Year, n = year_lags, default = NA), .names = "year_lag{.col}{i}")
  ) %>%
  ungroup()

# Number of lags you want to create for season
season_lags <- 7

# Columns for which you want to create season lag variables
season_lag_columns <- c("Temperature", "Precipitation", "Wind.Speed", "Humidity")

# Create season lag variables for the specified columns
lagged_cropasc <- year_lagged_cropasc %>%
  group_by(State, Crop, Season) %>%
  mutate(
    across(all_of(season_lag_columns), ~lag(., order_by = Season, n = season_lags, default = NA), .names = "season_lag{.col}{i}")
  ) %>%
  ungroup()
```

Figure 12a: Creation of Lag features

As seen in Figure 12a, the number of lags for the variable Year was set to 5 while the number of lags for the variable Season was set to 7. We observed that any further creation of lags beyond these numbers failed to improve the accuracies of the model significantly. An evaluation of the results Figures 12b and c indicate an increase in the performance accuracy for the RF model from 78.55% to 83.62% while the SVM model increases from 71.18% to 74.38%. Comparatively, the kappa statistics also rise from 76.4% to 82.01% for the RF model and from 68.86% to 72.03% for the SVM model.

```
Overall Statistics                          Overall Statistics

               Accuracy : 0.8362                          Accuracy : 0.7438
                 95% CI : (0.8205, 0.851)                   95% CI : (0.7255, 0.7614)
    No Information Rate : 0.1995             No Information Rate : 0.1995
    P-Value [Acc > NIR] : < 2.2e-16          P-Value [Acc > NIR] : < 2.2e-16

                  Kappa : 0.8201                             Kappa : 0.7203

 Mcnemar's Test P-Value : NA                 Mcnemar's Test P-Value : NA

Statistics by Class:                         Statistics by Class:

                     Class: Arhar Class: Bajra                    Class: Arhar Class: Baj
Sensitivity               1.00000     0.275000    Sensitivity          0.90323      0.062
Specificity               0.98388     1.000000    Specificity          0.99171      0.968
Pos Pred Value            0.81579     1.000000    Pos Pred Value       0.88608      0.065
Neg Pred Value            1.00000     0.974826    Neg Pred Value       0.99308      0.966
Prevalence                0.06664     0.034394    Prevalence           0.06664      0.034
```

Figure 12b: RF results Approach 3        Figure 12c: SVM results Approach 3

The F1 scores in Figure 12d also comparatively indicate better model performance overall in Approach 3 compared to approach 2. However, it is noteworthy that in this method too, RF surpasses SVM in terms of delivering superior model performance, achieving an accuracy rate of 83.62% compared to SVM's 74.38%.

| Class | RF_F1_Score_Approach3 | Class | SVM_F1_Score_Approach3 |
|---|---|---|---|
| Arhar | 0.6696429 | Arhar | 0.8714734 |
| Bajra | 0.4158416 | Bajra | 0.5187970 |
| Barley | 1.0000000 | Barley | 1.0000000 |
| Cotton | 0.9775281 | Cotton | 0.7571429 |
| Gram | 0.9503546 | Gram | 0.7019868 |
| Groundnut | 0.9596413 | Groundnut | 0.8160000 |
| Jowar | 0.9318182 | Jowar | 0.9230769 |
| Jute | 0.9462366 | Jute | 0.9896907 |
| Maize | 0.9331823 | Maize | 0.7667638 |
| Moong | 0.9680365 | Moong | 0.8558140 |
| Paddy | 1.0000000 | Paddy | 0.9906542 |
| Ragi | 0.9200000 | Ragi | 0.6972477 |
| Rapeseed and Mustard | 1.0000000 | Rapeseed and Mustard | 0.7638889 |
| Safflower | 0.6666667 | Safflower | 0.6666667 |
| Sesamum | 0.5810811 | Sesamum | 0.3913043 |
| Soyabean | 1.0000000 | Soyabean | 0.9800000 |
| Sunflower | 0.9841270 | Sunflower | 0.9508197 |
| Urad | 0.1563786 | Urad | 0.7633588 |
| Wheat | 0.9782609 | Wheat | 0.8333333 |

Figure 12d: Approach 3 F1 scores for RF and SVM models

Overall, as shown in the analysis of the three approaches, the accuracy of the developed models varies significantly, with the RF model consistently outperforming the SVM model across accuracy, kappa statistics and F1-Score performance metrics.

## 4. Discussion

### 4.1 Model performance

In all the three approaches, RF outperforms the SVM algorithm confirming its suitability in building crop recommendation models. It exhibits a remarkable accuracy of 99.96% in Approach 1 followed by 78.55 % in Approach 2 and 83.62% in Approach 3. This outcome aligns with the findings reported by Geetha et al (2020) and Lata and Chaudhary (2017) where RF emerged consistently as one of the most accurate algorithms for developing crop recommendation models among the other models employed. However, it diverges from the results obtained by Bondre and Mahagaonkar (2019) and Kumar et al. (2020) wherein the SVM algorithm outperformed the RF algorithm. This inconsistency in results can be attributed to the fact that both studies utilised datasets with limited features, as SVM tends to perform effectively with smaller datasets and fewer features (Varghese, 2018). Bondre and Mahagaonkar (2019) focused solely on soil parameters and crop yield while Kumar et al. (2020) only considered four parameters when constructing their recommendation model. The dataset utilised in our research included 15 features across weather, soil and market parameters, which may have introduced complexity for SVM in predicting optimal crops. On the other hand, as RF is robust to outliers and captures complexities in datasets with large features, it outperforms SVM for our recommendation model.

However, as depicted in the result section, although RF consistently outperforms SVM in all the three approaches, there is a significant difference in the accuracies noted for RF in all the approaches. Our result in the 10-fold cross validation approach corroborates with the work of Geetha et al (2020) and Lata and Chaudhary (2017). In both studies, the RF algorithm yielded impressive accuracy rates of 97% and 97.89% respectively, which closely aligns with the accuracy results of only Approach 1. To our knowledge, no study has revealed low accuracy as shown in Approach 2 (Time-Series Split) and Approach 3(Lag Variable) for RF. Doshi et al. (2018) who also trained their models on time-series data accumulated over the thirty-year period (from 1957-58 to 1986-87) gave an RF accuracy of 90.43%, which is much higher than results of our approach 2 and 3. The differing accuracy could relate to the consequence of testing the model on completely unseen data (year 2014). Unlike in Doshi et al's (2018) work where the train and test split was done randomly, we tested the models on unseen or new data, producing lower RF accuracy of 78.55% and 71.18% for SVM compared to the results of Doshi et al's (2018).

Furthermore, Approach 2 introduced a chronological order to the dataset to adapt to temporal changes but displayed significantly reduced accuracy and performance. This indicates the need for more advanced evaluations for models to properly adapt to new, previously unseen data. In other words, models need to be generalised to new data so that it can make accurate predictions when used (Wang et al., 2022; Volpi et al., 2018). To this end, the Lag variables approach (Approach 3) was introduced and lag features were incorporated to account for advanced temporal dependencies and seasonality, which enhanced the models' predictive capabilities compared to Approach 2. However, this could not achieve high accuracy compared to Approach 1. Overall, although Approach 3 has lower accuracy and kappa statistics than Approach 1, we argue that it still appears to be the most promising of the three approaches. It tests the model on completely unseen data and addresses temporal dependencies, seasonality and historical patterns through lag variables, resulting in improved model performance compared to Approach 2.

### 4.2 Input parameters

Regarding input parameters, Sita et al (2023) state that environmental factors like weather conditions are necessary to build an efficient crop recommendation model. However, a careful look at the variable importance graphs of all three RF approaches reveal that the economic factors - Operational Cost, Fixed Cost and MSP are consistently placed at the top (Figures 13a, b and c), suggesting that these variables hold greater significance and exert more influence when it comes to making recommendations. This discovery challenges the beliefs of previous researchers like Jain and Ramesh (2020) and Patel and Patel (2020) who placed higher importance on environmental variables to predict the optimum crop.

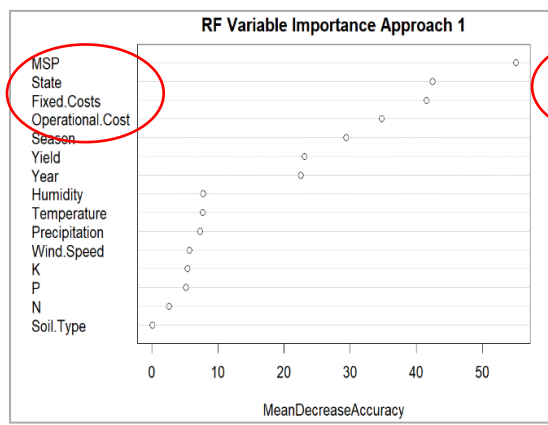
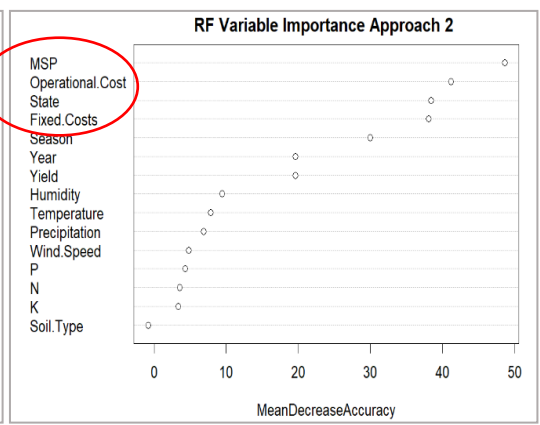

Figure 13a: RF variable importance Approach 1

Figure 13b: RF variable importance Approach 2

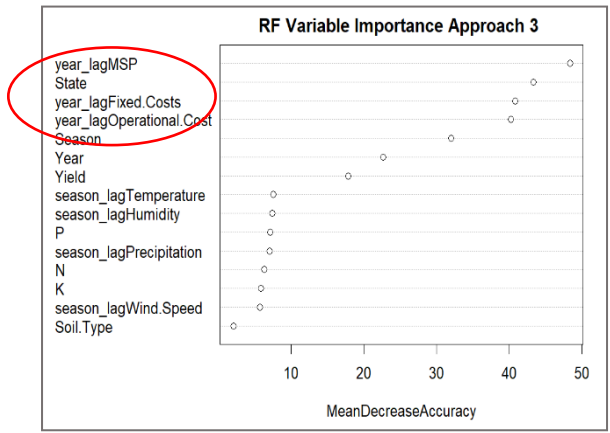

Figure 13c: RF variable importance Approach 3

### 4.3 Performance evaluation metrics

Many researchers including Kapoor and Verma (2017), Bondre and Mahagaonkar (2019) and Priyadarshini et al (2021) have traditionally relied solely on accuracy metrics to determine the most suitable crop recommendation model. Our study takes a more comprehensive approach by also incorporating Kappa statistics and F1 score for evaluating the developed models. The findings reported show that Kappa statistics is consistently lower than accuracy values for all the developed models across all the three approaches. For instance, in RF Approach 2, the accuracy stands at 78.55% whereas the Kappa score is 76.4%. This observation implies that while the accuracy may appear higher, the same model will give lower values of Cohen's kappa when dealing with unbalanced datasets (Widmann, 2020). To this end, it is quite likely that if researchers (e.g., Kapoor and Verma, 2017; Bondre and Mahagaonkar, 2019) had incorporated Kappa statistics as well as a measure of performance evaluation, the accuracy of their developed models could have been reduced, as Kappa statistic provides a more balanced and reliable representation of model accuracy when adjusted for such class imbalances. Additionally, while researchers like Arooj et al.(2017) have applied precision (accuracy of positive crop recommendations) and recall (model's ability to identify beneficial crops in relevant situations) independently as performance evaluation metrics. Our study goes beyond the use of F1 score and considers the trade-off between precision and recall, making it a useful single metric to evaluate the overall performance of the crop recommendation model.

### 5. Conclusion

This work proposed comprehensive crop recommendation models to help farmers in India optimise their crop selection decisions and reduce risk of crop failure while maximising profitability. Currently, most existing crop recommendation systems primarily rely on just environmental factors, cover limited geographical areas and recommend only a handful of crops thus creating a research gap. The model incorporates both environmental and economic parameters and 19 crop varieties across 15 different states.

Two supervised machine-learning models (Random Forest and Support Vector Machines) were developed and evaluated using three approaches – 10-fold Cross Validation Approach (Approach 1), Time-series split approach (Approach 2) and Lag variables approach (Approach 3). The models' performance were evaluated using accuracy, Kappa statistics and F1 scores. Analysis reveals that the 10-fold cross validation approach produced exceptionally high accuracy (RF: 99.96%, SVM: 94.71%), raising concerns of overfitting. However, the introduction of temporal order, which aligns more with real-world scenarios, reduces the model performance (RF: 78.55%, SVM: 71.18%) in the Time-series

Split approach. To further increase the model accuracy while maintaining the temporal order, the Lag Variables approach was employed, which resulted in improved performance (RF: 83.62%, SVM: 74.38%) compared to the 10-fold cross validation approach.

The models in the Time-series Split Approach and Lag Variable Approach offer practical insights by handling temporal dependencies and enhancing its adaptability to changing agricultural conditions over time. To this end, the study concludes that the Random Forest model developed based on the Lag Variables is the most preferred algorithm for optimal crop recommendation in the Indian context. The model maintains the temporal order of the dataset and captures fluctuations through lag variables resulting in more realistic crop recommendations. However, future work should employ advanced approaches to explore complex pattern recognition and improve prediction accuracy. Additionally, combining multiple algorithms or hybrid models that integrate traditional time series methods with machine learning approaches can yield improved performance in crop recommendation.

Furthermore, the study demonstrates the importance of considering both economic and environmental factors as input parameters for developing crop recommendation models. However, although these parameters are crucial and offer market considerations in the recommendation model, other important features need to be considered to develop robust recommendation models. Features like market demand, market supply, retail prices and return on investment can provide comprehensive insights and cater to the diversity of Indian agricultural practices.

**Data availability** – The data that support the findings of this study are openly available from three sources. The data: Kaggle (https://www.kaggle.com/datasets/vihith12/crop-yield_recommendation-dataset) for environmental parameters and the India Directorate of Economics and Statistics (https://eands.dacnet.nic.in/Cost_of_Cultivation.htm) and Farmer's Portal for economic parameters of cost and price (https://farmer.gov.in/mspstatements.aspx).

Crop Recommendation with Machine Learninglearning models for decision support system. *Journal of Emerging Technologies and Innovative Research (JETIR)*.

Pandian, S. (2023) K-fold cross validation technique and its essentials. *Analytics Vidhya.* Available at: https://www.analyticsvidhya.com/blog/2022/02/k-fold-cross-validation-technique-and-its-essentials/ [Accessed: 23-06-2023]

Patel, K., & Patel, H. B. (2020). A state-of-the-art survey on recommendation system and prospective extensions. *Computers and Electronics in Agriculture*, *178*, 105779.

Priyadharshini, A., Chakraborty, S., Kumar, A., & Pooniwala, O. R. (2021, April). Intelligent crop recommendation system using machine learning. In *2021 5th international conference on computing methodologies and communication (ICCMC)* (pp. 843-848). IEEE.

Pudumalar, S., Ramanujam, E., Rajashree, R. H., Kavya, C., Kiruthika, T., & Nisha, J. (2017, January). Crop recommendation system for precision agriculture. In *2016 eighth international conference on advanced computing (ICoAC)* (pp. 32-36). IEEE.

Rani, S., Mishra, A. K., Kataria, A., Mallik, S., & Qin, H. (2023). Machine learning-based optimal crop selection system in smart agriculture. *Scientific Reports*, *13*(1), 15997.

Reddy, D. J., & Kumar, M. R. (2021, May). Crop yield prediction using machine learning algorithm. In *2021 5th International Conference on Intelligent Computing and Control Systems (ICICCS)* (pp. 1466-1470). IEEE.

Sahoo, K., Samal, A. K., Pramanik, J., & Pani, S. K. (2019). Exploratory data analysis using Python. *International Journal of Innovative Technology and Exploring Engineering (IJITEE)*, *8*(12), 2019.

Schafer, J. L., & Graham, J. W. (2002). Missing data: our view of the state of the art. *Psychological methods*, *7*(2), 147.

Shmueli, B. (2021). Multi-Class Metrics Made Simple, Part III: the Kappa Score (aka Cohen's Kappa Coefficient). *Medium*. Available at: https://towardsdatascience.com/multi-class-metrics-made-simple-the-kappa-score-aka-cohens-kappa-coefficient-bdea137af09c [Accessed: 09-09-2023].

Shoaib, M., Shah, B., Ei-Sappagh, S., Ali, A., Alenezi, F., Hussain, T., & Ali, F. (2023). An advanced deep learning models-based plant disease detection: A review of recent research. *Frontiers in Plant Science*, *14*, 1158933.

Shumway, R., & Stoffer, D. (2019). *Time series: a data analysis approach using R*. Chapman and Hall/CRC.

Talagala, T. S., Hyndman, R. J., & Athanasopoulos, G. (2018). Meta-learning how to forecast time series. *Monash Econometrics and Business Statistics Working Papers*, *6*(18), 16.

Tanasa, D., & Trousse, B. (2004). Advanced data preprocessing for intersites web usage mining. *IEEE Intelligent Systems*, *19*(2), 59-65.

Varghese, D. (2018). Comparative study on classic machine learning algorithms. *Medium.* Available at : https://towardsdatascience.com/comparative-study-on-classic-machine-learning-algorithms-24f9ff6ab222 [Accessed:16-06-2023]

Volpi, R., Namkoong, H., Sener, O., Duchi, J. C., Murino, V., & Savarese, S. (2018). Generalizing to unseen domains via adversarial data augmentation. *Advances in neural information processing systems*, *31*.

Wang, J., Lan, C., Liu, C., Ouyang, Y., Qin, T., Lu, W., ... & Yu, P. (2022). Generalizing to unseen domains: A survey on domain generalization. *IEEE Transactions on Knowledge and Data Engineering*.
22